\pdfoutput=1

\documentclass[11pt]{article}

\usepackage{ACL}

\usepackage{times}
\usepackage{latexsym}
\usepackage[T1]{fontenc}
\usepackage{enumitem}
\usepackage{multirow}
\usepackage{url} 

\usepackage{booktabs}
\usepackage[utf8]{inputenc}
\usepackage{hyperref}

\usepackage{graphicx}
\usepackage{caption}
\usepackage{subcaption}
\usepackage[acronym]{glossaries}
\glsdisablehyper
\newacronym{LM}{LM}{Language Model}
\newacronym{LMs}{LMs}{Language Models}
\newacronym{LLM}{LLM}{Large Language Model}
\newacronym{LLMs}{LLMs}{Large Language Models}
\newacronym{QA}{QA}{Question Answering}
\newacronym{BPE}{BPE}{Byte-Pair Encoding}
\newacronym{BBPE}{BBPE}{Byte-Level BPE}
\newacronym{SP}{SP}{SentencePiece}
\newacronym{HF}{HF}{Hugging Face}

\usepackage[capitalize]{cleveref}

\usepackage{microtype}
\usepackage{footnote}
\usepackage{inconsolata}
\makeatletter
\newcommand{\printfnsymbol}[1]{%
  \textsuperscript{\@fnsymbol{#1}}%
}
\makeatother

\title{Tokenizer Choice For LLM Training: Negligible or Crucial?}

\author{
   Mehdi Ali\textsuperscript{1,2} $^\dagger$, Michael Fromm\textsuperscript{1,2} $^\dagger$, Klaudia Thellmann\textsuperscript{3} $^\dagger$ \\
   Richard Rutmann\textsuperscript{1,2}, Max Lübbering\textsuperscript{1,2}, Johannes Leveling\textsuperscript{1}, Katrin Klug\textsuperscript{1}, Jan Ebert\textsuperscript{4},  \\
   Niclas Doll\textsuperscript{1}, Jasper Schulze Buschhoff\textsuperscript{1}, Charvi Jain\textsuperscript{1,2}, Alexander Arno Weber\textsuperscript{1,2}, \\
   Lena Jurkschat\textsuperscript{3}, Hammam Abdelwahab\textsuperscript{1}  
   Chelsea John\textsuperscript{4}, Pedro Ortiz Suarez\textsuperscript{5}, Malte Ostendorff\textsuperscript{5} \\ Samuel Weinbach\textsuperscript{6}, Rafet Sifa\textsuperscript{1}, Stefan Kesselheim\textsuperscript{4}, Nicolas Flores-Herr\textsuperscript{1}   \\   \\ 
  \textsuperscript{1}Fraunhofer IAIS, \textsuperscript{2}Lamarr Institute, \textsuperscript{3}TU-Dresden, \textsuperscript{4}FZ Jülich, \textsuperscript{5}DFKI, \textsuperscript{6}Aleph Alpha
   \Thanks{\textdagger Equal contribution.}
}

\begin{document}
\maketitle
\begin{abstract}
The recent success of \gls{LLMs} has been predominantly driven by curating the training dataset composition, scaling of model architectures and dataset sizes and advancements in pretraining objectives, leaving tokenizer influence as a blind spot.
Shedding light on this underexplored area, we conduct a comprehensive study on the influence of tokenizer choice on LLM downstream performance by training 24 mono- and multilingual LLMs at a 2.6\,B parameter scale, ablating different tokenizer algorithms and parameterizations. Our studies highlight that the tokenizer choice can significantly impact the model's downstream performance and training costs. 
In particular, we find that the common tokenizer evaluation metrics \textit{fertility} and \textit{parity} are not always predictive of model downstream performance, rendering these metrics a questionable proxy for the model's downstream performance. Furthermore, we show that multilingual tokenizers trained on the five most frequent European languages require vocabulary size increases of factor three in comparison to English. 
While English-centric tokenizers have been applied to the training of multi-lingual \gls{LLMs} in the past, we find that this approach results in a severe downstream performance degradation and additional training costs of up to 68\%, due to an inefficient tokenization vocabulary.
\end{abstract}

\section{Introduction}

\gls{LLMs} have shown impressive capabilities in many downstream tasks in a zero/few-shot setting such as summarization, reading comprehension, translation, and commonsense reasoning~\cite{DBLP:conf/nips/BrownMRSKDNSSAA20,DBLP:journals/corr/abs-2307-09288}.
To train a LLM, the currently established approach is to employ a tokenizer that splits the training documents into tokens where a token represents a word~\cite{bengio2000neural}, a sub-word~\cite{schuster2012japanese,sennrich2015neural,Wang_Cho_Gu_2020}, or a single character~\cite{gao-etal-2020-character}, and each token is represented in the model by an embedding vector that can be further processed.

The quality of a tokenizer can be assessed \textit{intrinsically} and \textit{extrinsically}.  An intrinsic evaluation solely addresses the characteristics of tokenizers and their generated output in isolation, whereas the extrinsic evaluation measures the impact of the tokenizer on a downstream component, e.g., the \gls{LLM}.

While many different tokenization approaches have been proposed, ranging from character-based to word-based methods, the potential impact of different tokenizers is underexplored w.r.t. \gls{LLMs}, especially in the context of multilingual \gls{LLMs}.
Recent work proposed by~\citet{petrov2023language} demonstrates that carelessly designed tokenizers applied to the training of multilingual \gls{LLMs} result in severe inequalities and limitations across languages. 
Text passages translated into different languages resulted in tokenized sequences that differ in length up to a factor of 15, affecting inference costs and latency during inference.
Furthermore, it is known that the learning of long-range dependencies~\cite{DBLP:conf/nips/VaswaniSPUJGKP17}, is an essential property for effectively learning transformer-based \gls{LLMs}.
Given a fixed sequence length, learning to relate words far apart in the input text is impossible for languages whose text is excessively fragmented by the tokenizer.

Despite the importance of tokenizers and the potentially severe impact of poorly performing tokenizers, there exists no extensive study so far that holistically investigates the intrinsic and extrinsic tokenizer performance in a monolingual and multilingual setting with a focus on decoder-only models, which represent the backbone of current \gls{LLMs}.

In this work, we address this gap and conduct an extensive study in which we measure the impact of the tokenizer on the model performance.
In particular, we make the following contributions:

\begin{itemize}
    \item We conduct a study investigating the intrinsic tokenizer performance.
    \item We conduct a study investigating the extrinsic tokenizer performance, i.e., the impact of the tokenizer on the model's downstream performance.
    \item We investigate whether a correlation between the intrinsic and the extrinsic tokenizer performance exists.
\end{itemize}

\section{Related Work}

This section provides an overview of tokenization algorithms and their usage in encoder- and decoder-only transformer models.

\subsection{Tokenization Approaches}
\paragraph{Word Tokenization.}
The most basic tokenization approach is the splitting of sequences based on white spaces and considering each word as a token~\cite{bengio2000neural}.

\paragraph{Subword tokenization.}
This class of algorithms subsumes all data-driven tokenization approaches which can decompose words into subwords/multiple tokens and currently represent the established tokenization approach upon which \gls{LLMs} rely~\cite{kudo2018sentencepiece,petrov2023language}.
Because subword tokenizers decompose words into subwords, they can process out-of-vocabulary words by merging subwords from the vocabulary~\cite{kudo2018sentencepiece}.
Examples of popular subword tokenizers are WordPiece~\cite{schuster2012japanese}, BPE~\cite{Gage1994ANA,sennrich2015neural}, \gls{BBPE}~\cite{Wang_Cho_Gu_2020}, and Unigram~\cite{kudo-2018-subword}.

\paragraph{Character Tokenization.}

Tokenization can also be performed on a character level or based on UTF-8 bytes.
However, this results in an increased sequence length, which becomes computationally expensive in the transformer architecture, the current predominated architecture for \gls{LLMs} due to the quadratic complexity of the self-attention layer in the sequence length~\cite{DBLP:conf/nips/VaswaniSPUJGKP17}.
Though, several approaches have been proposed to address this limitation \cite{gao-etal-2020-character, tay2021charformer, xue2022byt5, clark-etal-2022-canine, yu2023megabyte}.

\subsection{Tokenizers in Transformers Models}\label{subsec:rel_tokenizer_transformer}

\paragraph{Tokenizers in Encoder Models}  Most research on tokenization has been conducted on encoder models. 
\citet{rust-etal-2021-good} investigated whether the tokenizer choice impacts the downstream performance of multi- and monolingual BERT \cite{devlin2018bert} models. 
\citet{zhang-etal-2022-robust} showed that better machine translation performance is often obtained when languages are equally sampled during the tokenizer training.
\citet{10.1145/3578707} trained several medium-sized language models for Turkish and suggested that different subword tokenizers perform roughly equivalent, whereas word- and character-level tokenizers perform drastically worse on downstream tasks.
Finally,~\cite{chirkova2022codebpe} analyzed the effect of employing different tokenizations on code-related tasks and demonstrated that carefully configured tokenizers could reduce average sequence length up to 40\% or allow for small downstream performance improvements by up to 2\% at a lower compression rate.

\paragraph{Tokenizers in Decoder Models}\label{subsec:rel_tokenizer_decoder}

An overview of current mono- and multilingual \gls{LLMs} is provided in~\cite{lin2022few, shliazhko2022mgpt, DBLP:journals/corr/abs-2211-05100}. \citet{stollenwerk2023training} evaluated the intrinsic metrics of the GPT-SW3 \cite{ekgren2023gptsw3} tokenizer that focused on the Nordic languages.
As part of their work, \citet{shliazhko2022mgpt} ablated different tokenizer pre-processing approaches while keeping the tokenizer algorithm, the vocabulary size, and the employed implementation fixed. 
In none of the other major \gls{LLM} publications, the extrinsic tokenizer performance has been studied.

\section{Approach}

To investigate the tokenizer impact on the model performance, we conducted an extensive ablation study. In detail, we created dedicated datasets for the training  of the tokenizers and the models, trained BPE and Unigram tokenizers, and for each tokenizer we trained decoder-only models with a size of 2.6B parameters while keeping the remaining configuration (i.e., dataset and model hyper-parameters) fixed.
This allowed us to measure the tokenizer's impact on the model's downstream performance in isolation.

\subsection{Data}

While creating our tokenizer and model training datasets, we ensure that the mixture proportions of data domains (Wikipedia, books, web text) follow the same distribution to avoid a domain shift between tokenizers training and model training. 
We created \emph{two datasets} with 70B words where one of the datasets is monolingual, containing English documents, and the second is a multilingual dataset comprised of English, German, French, Italian, and Spanish documents. 
Our datasets are filtered and deduplicated and consist of web-crawled data (80\%) and curated data (20\%), comparable to related datasets used to train \gls{LLMs}. 
In the multilingual dataset, the amount of web-crawled data is equally distributed across languages in terms of number of words.
Further details about our data pipeline and the data composition are described in \cref{appendix:corpora}.

\subsection{Tokenizer}

Our studies rely on the two established tokenization algorithms, BPE and Unigram, and their implementation in the \textit{Huggingface tokenizer} library \cite{Moi_HuggingFace_s_Tokenizers_2023}  and the \textit{SentencePiece} library \cite{kudo2018sentencepiece}.
We considered both libraries in order to investigate the effect of differences in the pre-and post-processing steps and potential differences in the implementations.
Due to missing pre-processing options for Huggingface's Unigram implementation, which causes a large discrepancy in the resulting vocabulary compared to SentencePiece's implementation of Unigram, we omitted the training of Unigram tokenizers based on Huggingface.
Overall, we trained 24 different tokenizers, where one-half of the tokenizers were monolingual English tokenizers, and the other half of the tokenizers were multilingual tokenizers.
Besides the tokenizer algorithm, language composition, and employed tokenizer library, we also varied the vocabulary size.
Concrete tokenizer configurations are described in the \cref{appendix:tokenizer}.

\subsection{Models}

To measure the impact of our trained tokenizers on the model downstream performance, we trained one model for each tokenizer.
In particular, for each of our 24 trained tokenizers, we trained a 2.6B transformer-based decoder-only model on up to 52B tokens following the scaling law proposed by~\cite{NEURIPS2022_c1e2faff}.
Additionally, serving as baselines, we trained a monolingual and a multilingual model using the pre-trained GPT-2 tokenizer~\cite{radford2018improving}. All models have been trained based on the causal language modeling training objective.

\subsection{Evaluation}

To assess the impact of the tokenizers on the model downstream performance, we first performed an intrinsic tokenizer evaluation, followed by an extrinsic evaluation, and finally, we investigated whether a correlation between both evaluation approaches is given.

The intrinsic evaluation aims to assess the generated output of tokenizers based on \textit{fertility} and \textit{parity}.
Furthermore, the tokenizer's vocabulary overlap with other tokenizers is computed.
The intrinsic evaluation does not assess the impact of tokenizers on the model performance.

Fertility, the most common metric to evaluate a tokenizer's performance \cite{DBLP:journals/corr/abs-2211-05100,stollenwerk2023training, rust-etal-2021-good}, is defined as the average number of tokens that are required to represent a word or document. 
For a tokenizer $T$ and dataset $A$, the fertility can be calculated as the number of tokens in $A$ (when $T$ is applied) divided by the number of words in $A$. 
We calculate the fertility on a held-out set (10,000 documents), which was not used for the tokenizer training. 
For calculating the words of a document, we used whitespace splitting.
Higher fertility scores correspond to weaker compression capabilities of the tokenizer.

Parity~\cite{petrov2023language}, which has been recently proposed, assesses how fairly a tokenizer treats equivalent sentences in different languages.
A tokenizer $T$ achieves parity for language $A$ with respect to language $B$ if $\frac{|T(s_A)|}{|T(s_B)|} \approx 1$, where $s_A$ and $s_B$ denote the sets of all sentences in the corpora of languages $A$ and $B$, respectively, and the ratio $\frac{|T(s_A)|}{|T(s_B)|}$ is defined as premium.
We use the FLORES-200 \cite{10.1162/tacl_a_00474} parallel corpus, consisting of the same sentences human-translated into 200 languages.
We calculate the parity values for each tokenizer and the four non-English languages with respect to English (see \cref{fig:parity_de_it_es_fr} for an overview).

The extrinsic evaluation aims to explicitly assess the impact of a tokenizer on the model's downstream performance.
We selected a comprehensive set of downstream tasks (see \cref{experimental-setup}) to measure the downstream performance. 

Additionally, we computed the impact of a tokenizer on the average computational costs of a given model per word during training.
The computational costs during training for one step including the forward and the backward pass can be estimated by
\begin{equation}
    C = 96Bslh^2\left(1 + \dfrac{s}{6h} + \dfrac{V}{16lh} \right), \label{eq:training_costs}
\end{equation}
given a model with batch size $B$, sequence length $s$, $l$ layers, hidden size $h$ and vocabulary size $V$ \cite{10.1145/3458817.3476209}. 
The costs per token can be derived by $C_{\text{token}} = C/Bs$ and the average costs per word by $C_{\text{word}} = C_{\text{token}} \times \text{fertility}$.
The Results are discussed in \cref{subsec:costs}.

\section{Intrinsic Tokenizer Evaluation}\label{sec:instinsic_tok_eval}

In our intrinsic evaluation, we first compare the fertility and parity of the trained tokenizers (Section~\ref{subsec:fertility}) and subsequently the overlap of their vocabularies (Section~\ref{subsec:vocab_overlap}).

\subsection{Fertility \& Parity}\label{subsec:fertility}
\begin{figure*}
     \centering
     \begin{subfigure}[b]{0.495\textwidth}
         \centering
         \includegraphics[width=\textwidth]{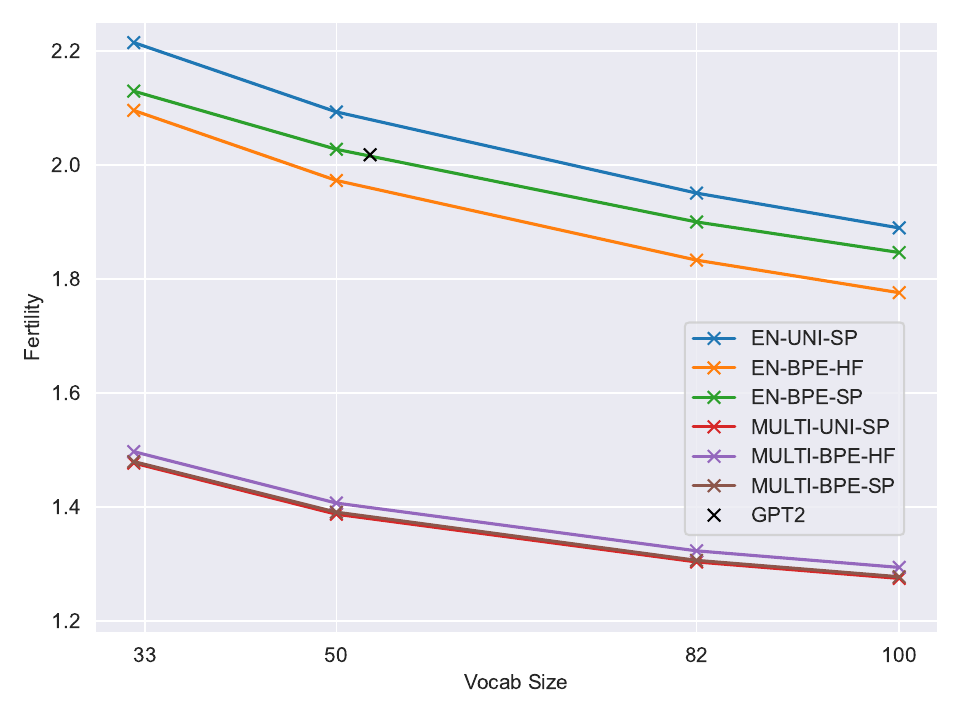}
         \caption{Non-English, multilingual documents}
         \label{fig:fertility_de_it_es_fr}
     \end{subfigure}
     \hfill
        \begin{subfigure}[b]{0.495\textwidth}
         \centering
         \includegraphics[width=\textwidth]{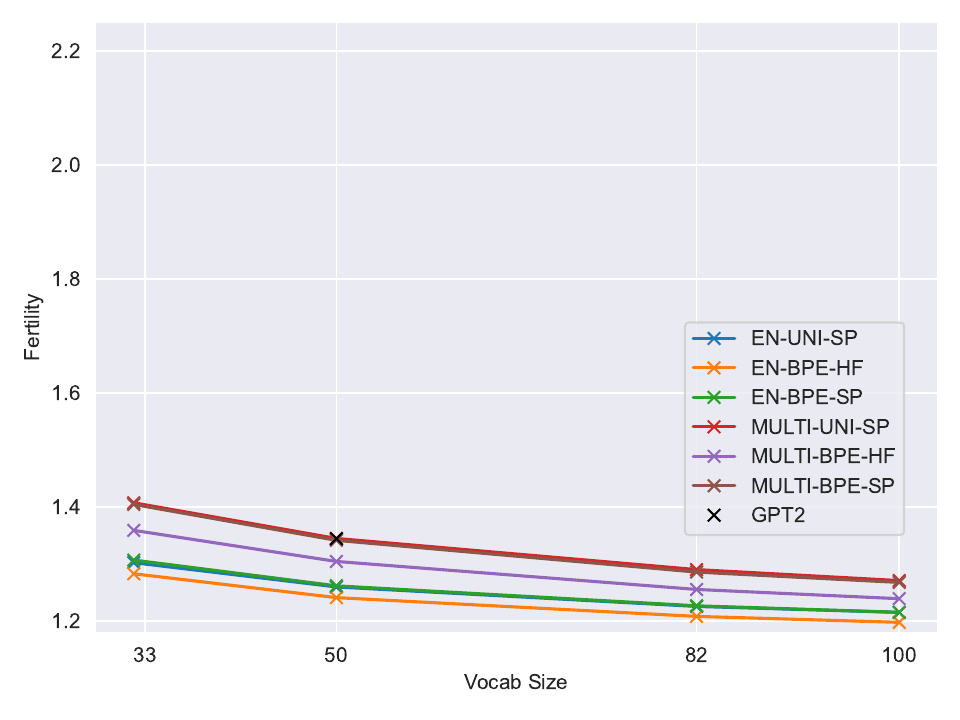}
         \caption{English documents}
         \label{fig:fertility_english_only}
     \end{subfigure}
        \caption{Comparison of fertility scores between mono- and  multilingual tokenizers applied to (a) Non-English, multilingual documents and (b) English documents.}
        \label{fig:fertility}
\end{figure*}

\begin{figure}
         \centering
         \includegraphics[width=0.495\textwidth]{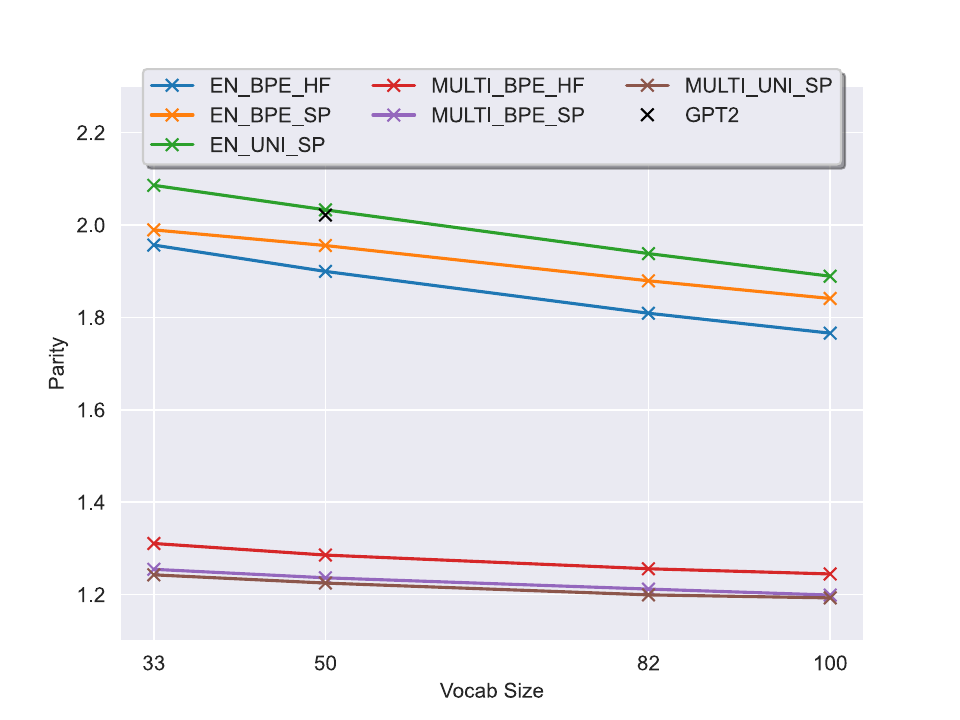}
         \caption{Comparison of parity scores between monolingual (English) tokenizer and multilingual tokenizers applied multi-lingual documents.}
         \label{fig:parity_de_it_es_fr}
\end{figure}

Applying the described fertility and parity evaluation to the mono-/multilingual tokenizers, our analysis highlights the following two major aspects, as visualized in~\cref{fig:fertility}~and~\cref{fig:parity_de_it_es_fr}.

Firstly, it can be observed that applying a monolingual tokenizer to multilingual data results in significantly higher fertility and parity scores (see \cref{fig:fertility_de_it_es_fr} and \cref{fig:parity_de_it_es_fr}). While multilingual tokenizers have lower fertility than monolingual English tokenizers on all non-English documents by a large margin, they are only slightly worse on tokenizing English documents, as shown in \cref{fig:fertility_english_only}. 

Secondly, with increasing vocabulary size, fertility and parity reduce in all cases, which can be explained by the tokenizer requiring fewer sub-word tokens when tokenizing text given a larger vocabulary.
However, it can be observed that for monolingual English tokenizers, the fertility is less dependent on the vocabulary when tokenizing English documents, implying that 33k might be a sufficiently large vocabulary.

\subsection{Vocabulary Overlap}\label{subsec:vocab_overlap}

To analyze the tokenizer similarity, we calculated the vocabulary overlap. Particularly, we assess Huggingface's and SentencePiece's BPE implementations, as depicted in \cref{tab:vocab_overlap}.

The overlap is roughly constant across different vocabulary sizes, and the total overlap tends to be rather low, despite being the identical algorithm only implemented by two different libraries. Consequently, the tokenizers produce different tokenized sequences, possibly affecting model training and downstream performance.
Investigating the underlying reasons, the low overlap might be attributed to different configuration and pre-processing options in these libraries. Due to the larger thesaurus in multilingual documents, the overlap for the multilingual tokenizer is lower than for the English tokenizers.

\begin{table}
    \centering
    \begin{tabular}{lcccc}
    \toprule
         & 33k & 50k & 82k & 100k\\
        \midrule
         English & 0.77  & 0.76  & 0.74  & 0.74 \\
         Multilingual & 0.62  & 0.62  & 0.62  & 0.61 \\
         \bottomrule
    \end{tabular}
    \caption{Vocabulary overlap between the HuggingFace and SentencePiece BPE tokenizer for different vocab sizes.}
    \label{tab:vocab_overlap}
\end{table}

\begin{figure*}[ht]
     \centering
     \begin{subfigure}[b]{0.325\textwidth}
         \centering
         \includegraphics[width=\textwidth]{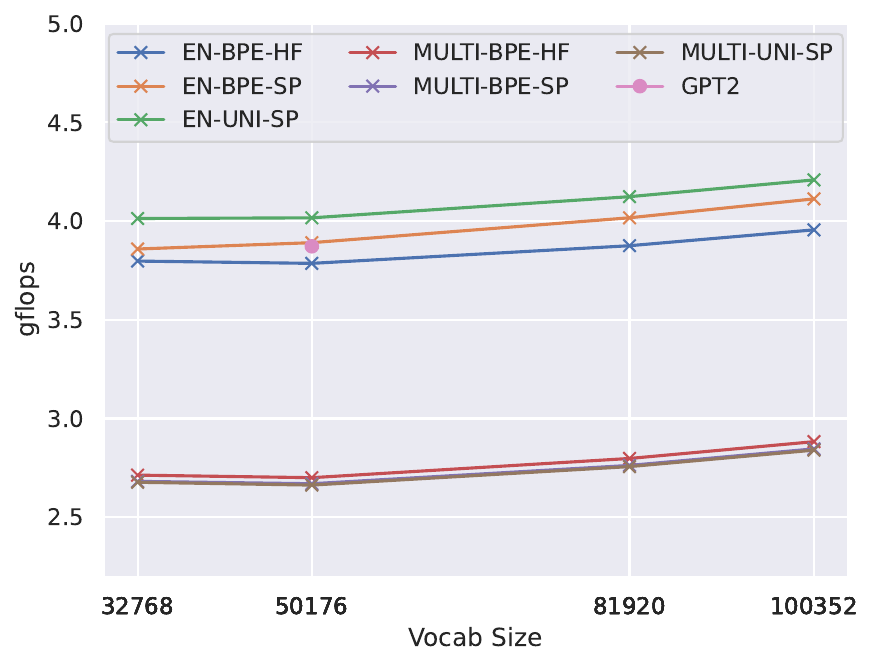}
         \caption{Non-English documents}
         \label{fig:costs_training_multi}
     \end{subfigure}
     \hfill
          \begin{subfigure}[b]{0.325\textwidth}
         \centering
         \includegraphics[width=\textwidth]{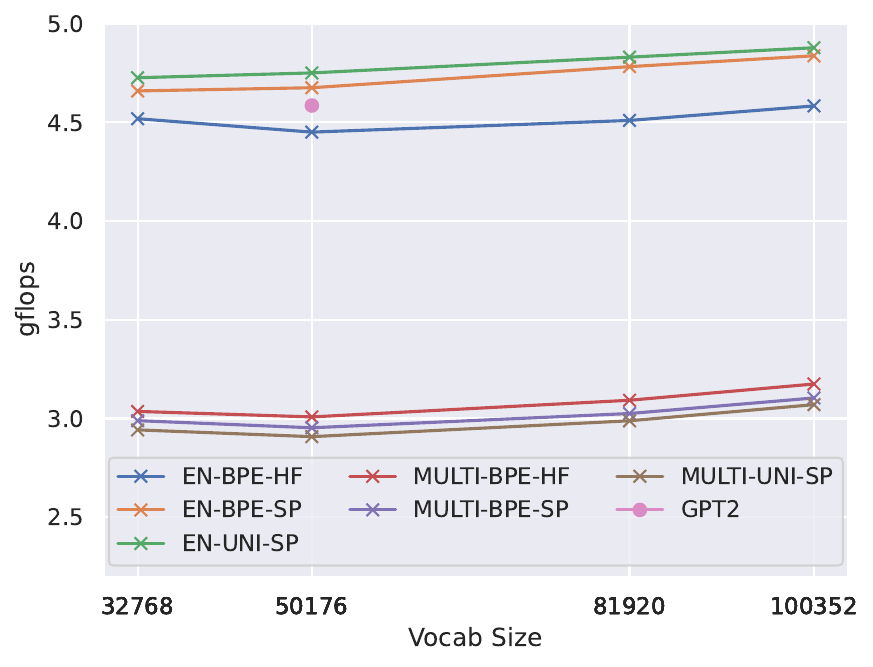}
         \caption{German documents}
         \label{fig:costs_training_de}
     \end{subfigure}
    \hfill
     \begin{subfigure}[b]{0.325\textwidth}
         \centering
         \includegraphics[width=\textwidth]{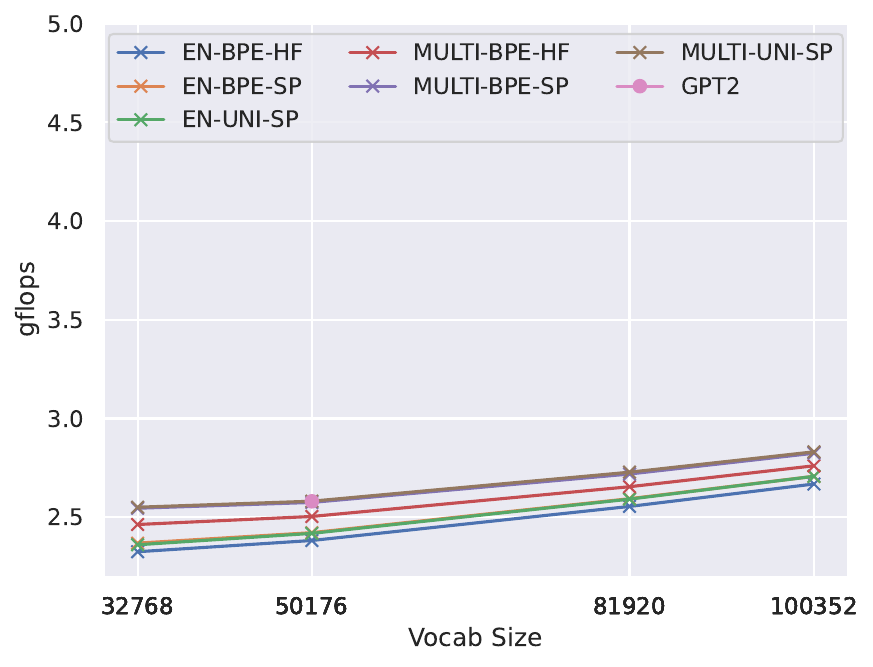}
         \caption{English documents}
         \label{fig:costs_training_en}
     \end{subfigure}

        \caption{Average compute (GFLOPS) required to process a single word within (a) multilingual, (b) English, and (c) German documents within a full \textbf{training} pass (including the backward pass).}
        \label{fig:costs_training}
\end{figure*}

\section{Extrinsic Tokenizer Evaluation}\label{sec:extinsic_tok_eval}
In the following, we describe the results of our extrinsic evaluation of tokenizers.
\cref{experimental-setup} describes the experimental setup, \cref{downstream_performance} presents the downstream performance of the trained models based on the investigated tokenizers, and \cref{subsec:costs} analyzes the computational costs associated with each tokenizer when employed in a specific model.

\subsection{Experimental Setup}\label{experimental-setup}

To assess the impact of the tokenizers on the model downstream performance, we trained a decoder-only transformer model of size 2.6\,B for each tokenizer.
We trained our models for 52.6\,B tokens following the scaling laws proposed by~\citet{DBLP:journals/corr/abs-2203-15556}, based on the causal language modeling training objective. 
The hyper-parameters are described in \cref{tab:hyperparams} in the \cref{appendix:architecture_params}. 
We evaluated our models in zero-shot settings on a wide range of mono- and multilingual tasks:
\begin{itemize}
    \item Natural language inference: XNLI~\cite{DBLP:conf/emnlp/ConneauRLWBSS18}, MNLI ~\cite{williams-etal-2018-broad}, RTE~\cite{wang-etal-2018-glue}, WNLI~\cite{levesque2012winograd}, CB~\cite{de2019commitmentbank}
    \item Question answering: 
    X-CSQA~\cite{DBLP:journals/corr/cs-CL-0108005}, XStoryCloze~\cite{lin2022few}, PubMedQA~\cite{jin-etal-2019-pubmedqa}
    \item Reading comprehension: BoolQ~\cite{DBLP:conf/naacl/ClarkLCK0T19}), LAMBADA~\cite{DBLP:conf/acl/PapernoKLPBPBBF16}, RACE \cite{lai-etal-2017-race}, MRPC \cite{dolan-brockett-2005-automatically}.
    \item Commonsense reasoning: HellaSwag~\cite{DBLP:conf/acl/ZellersHBFC19}, WinoGrande~\cite{DBLP:conf/aaai/SakaguchiBBC20}, ARC~\cite{DBLP:journals/corr/abs-1803-05457}, XCOPA~\cite{ponti-etal-2020-xcopa}, XCDOAH~\cite{DBLP:journals/corr/cs-CL-0108005}, WSC~\cite{levesque2012winograd}, COPA~\cite{roemmele2011choice}
    \item Classification: 
    PAWS-X~\cite{yang-etal-2019-paws}, GNAD10~\cite{Schabus2017}, SST~\cite{socher-etal-2013-recursive}, WIC~\cite{pilehvar-camacho-collados-2019-wic}, PIQA~\cite{bisk2020piqa}
\end{itemize}

\cref{tab:task_overview} provides an overview of the number of tasks for each category and language.

\begin{table}
\centering
\begin{tabular}{lccccc}
\toprule
 Task & EN & DE & FR & ES & IT   \\
\midrule
NLI  & 6 & 1 & 1 & 1 & 0  \\
QA & 3 & 2 & 2 & 3 & 2  \\
RC & 3 & 1 & 1 & 1 & 1  \\
CR & 7 & 0 & 1 & 0 & 1  \\
CL & 3 & 1 & 0 & 1 & 0  \\
\midrule
 & 22 & 5 & 4 & 6 & 4 \\
\bottomrule
\end{tabular}
\caption{Overview of the number of evaluation tasks for each  language and the categories of Natural language inference (NLI), Reading comprehension (RC), Question answering (QA), Commonsense reasoning (CR) and Classification (CL).}
\label{tab:task_overview}
\end{table}

\subsection{Downstream Performance}\label{downstream_performance}

We split our analysis of the downstream performance into several parts.

First, we discuss the overall results obtained for the investigated tokenizers, followed by presenting the impact of the tokenizer library (\cref{sec:tok_library}), the impact of the tokenizer algorithm (\cref{sec:tok_algorithm}), and the impact of the vocabulary size (\cref{sec:tok_vocabulary}).

We present both, aggregated results across all tasks (\cref{tab:average_accuracy}) and results for selected single tasks (\cref{tab:results_single_tasks}).
For the average performance across all tasks presented in \cref{tab:average_accuracy}, we computed weighted average to take into account the different number of tasks per language.
In particular, we computed for each language the mean across all tasks, and then computed the mean over all language-means.

\begin{table}
\centering
\begin{tabular}{llr}
\toprule
Model & EN & MULTI \\
\midrule
GPT-2-50 & 50.36 & 39.41 \\
\midrule
BPE-HF-33 & 49.13 &  40.52\\
BPE-HF-50 & 49.51 & 40.47 \\
BPE-HF-82 & 48.71 & 40.24\\
BPE-HF-100 & 49.54 & 40.48 \\
\midrule
BPE-SP-33 & \textbf{50.81} & 40.28 \\
BPE-SP-50 & 49.81 & 40.49\\
BPE-SP-82 & 48.99  & 41.21\\
BPE-SP-100 & 49.46 & \textbf{41.44}\\
\midrule
UNI-SP-33 & 50.28 & 40.30\\
UNI-SP-50 & 49.90 & 40.48\\
UNI-SP-82 & 49.65 & 41.20\\ 
UNI-SP-100 & 50.21 & 40.74\\
\midrule
\end{tabular}
\caption{Average accuracy of monolingual and multilingual tokenizers across all downstream tasks. Due to varying number of tasks per language, multi-lingual accuracies have been adjusted to each language contributing equally to the average.}
\label{tab:average_accuracy}
\end{table}

\begin{table}
\centering
\begin{tabular}{llccc}
    \toprule
    & Task & Min & Max & Rand. \\
    \midrule
    \multirow{4}{*}{\rotatebox{90}{EN}} 
    & ARC-Easy & 0.50  & 0.59 & 0.20 \\
    & HellaSwag & 0.34 & 0.41  &  0.25 \\
    & MRPC & 0.54 & 0.69 & 0.50 \\
    & PIQA & 0.67 & 0.72  & 0.50 \\
    \midrule
    \multirow{4}{*}{\rotatebox{90}{MULTI}} 
    & XNLI FR & 0.37 & 0.49  & 0.33 \\
    & XNLI EN & 0.49 & 0.52  & 0.33 \\
    & X-CODAH ES & 0.28 & 0.43  & 0.25 \\
    & 10kGNAD & 0.15 & 0.43 & 0.11 \\
    \bottomrule
\end{tabular}
\caption{Worst- and best-performing tokenizer for selected tasks and the random performance on this task.}
\label{tab:results_single_tasks}
\end{table}

\begin{table*}
\centering
\begin{tabular}{lccccccc}
\toprule
 & \multicolumn{6}{c}{MULTI} &  MONO \\ 
 \midrule
 Vocabulary & DE & FR & IT & ES & EN & AVG & EN \\
\midrule
 33  & \textbf{36.75} & 36.66 & 39.30 & 41.76  & 47.37 & 40.37 & 49.55 \\
 50 & 36.12 & 37.07 & 38.94 & 42.22 & 46.71  & 40.21 & \textbf{49.90} \\
 82 & 36.50 & 37.83 & 39.97 & 42.30 & \textbf{47.80} & 40.88 & 49.12  \\
 100 & 35.92 & \textbf{38.07} & \textbf{40.13} & \textbf{42.64} & 47.67 & \textbf{40.89} & 49.74  \\ 
\midrule\midrule
Algorithm and Library & DE  &  FR & IT & ES  & EN & AVG & EN \\
\midrule
 BPE-HF & 35.69 & 37.31 & 39.37 & 42.28 & 47.48 & 40.43 & 48.98 \\
 BPE-SP & \textbf{37.13} & 37.45 & \textbf{40.04} & 41.96 & \textbf{47.68} & \textbf{40.85} & 49.77 \\
 UNI-SP & 36.51 & \textbf{37.66} & 39.57 & \textbf{42.56} & 47.10 & 40.68 & \textbf{50.01} \\
\bottomrule
\end{tabular}
\caption{Impact of the vocabulary size (upper),  and tokenizer algorithm and library (lower), on the downstream performance. The accuracy scores are either averaged over the libraries and tokenizer algorithms (upper) or the different vocabulary sizes (lower).}
\label{tab:combined_tables}
\end{table*}

\paragraph{Monolingual Tokenizer} \cref{tab:average_accuracy} demonstrates that the BPE-SP-33 tokenizer, on average, is the best-performing tokenizer, followed by the GPT-2 tokenizer. Interestingly, SentencePiece's implementation of BPE with a vocabulary size of 33k has been used for LLaMA2 \cite{DBLP:journals/corr/abs-2307-09288}.
Aggregated metrics provide a reasonable overview of the overall performance. 
However, it does not express potentially large performance differences across tasks.
Therefore, we listed in \cref{tab:results_single_tasks} the obtained results for a list of selected tasks obtained by the best and worst performing tokenizer on this task.  
The results illustrate that the performance difference can be huge.
For instance, for ARC-Easy, a commonsense reasoning task, the gap between the best and worst tokenizer is 9\%. 

\paragraph{Multilingual Tokenizer}

\cref{tab:average_accuracy} shows that the BPE-SP-100 tokenizer is the best-performing tokenizer followed by the BPE-SP-82 tokenizer. 
Furthermore, \cref{tab:average_accuracy} demonstrates that the GPT-2 tokenizer performs poorly, implying that using a pre-trained GPT-2 tokenizer to pre-train and fine-tune multilingual models should be \textbf{omitted}.
The analysis of selected tasks (~\ref{tab:results_single_tasks}) reveals that for multilingual tokenizers, the performance difference between tasks can be huge.

\subsubsection{Impact of the Tokenizer Library}\label{sec:tok_library}
\cref{tab:combined_tables} demonstrates that BPE-SP, on average, outperforms BPE-HF in the monolingual and multilingual setting across all languages. 
The performance differences might be attributed to the differences in implementation details of the tokenizers' pre-and postprocessing, which could affect the vocabulary creation (see \cref{subsec:vocab_overlap}) and, consequently, the downstream performance.

\subsubsection{Impact of the Tokenizer Algorithm}\label{sec:tok_algorithm}
Furthermore, \cref{tab:combined_tables} shows that depending on the language, either the BPE or Unigram exhibits better performance.
It is noteworthy that the Germanic languages German and English benefit from the BPE algorithm, whereas the Romanic languages French and Spanish benefited from Unigram.
The experiments for Italian, a Romanic language as well, show a different pattern than the other two Romanic languages.

\subsubsection{Impact of the Tokenizer Vocabulary}\label{sec:tok_vocabulary}

Analyzing the impact of the vocabulary size revealed that in the monolingual English setting, the smaller/medium-sized, i.e., a vocabulary size of 33k/50k performs better (\cref{tab:combined_tables}) whereas in the multilingual setting, in all cases except for German, larger vocabulary sizes result in better downstream performance. 
Taking into account the results presented in \cref{tab:average_accuracy} showing that in the monolingual English setting, the best-performing tokenizer on average across all tasks had a vocabulary size of 33k and that the best-performing multilingual tokenizer had a vocabulary size of 100k additionally supports the observation that for the monolingual English setting a small vocabulary size is beneficial and for the multilingual setting a large vocabulary size is required.

\subsection{Computational Costs}\label{subsec:costs}

Given a fixed model, the computational costs depend on the vocabulary size and the fertility of the tokenizer, as defined in \cref{eq:training_costs}.

While larger vocabulary sizes introduce additional computational costs, they might also result in lower fertility scores and, therefore, lower overall computational costs for processing a set of documents, as discussed in \cref{sec:instinsic_tok_eval}.
However, our findings in \cref{fig:costs_training} show that increasing the vocabulary size from 50k to larger vocabulary sizes increases the computational costs in all cases. 
This highlights that the potentially lower fertility of larger vocabulary sizes cannot compensate for the additional costs introduced by the larger vocabulary size. 

Furthermore, we observe that the computational training costs for multilingual documents are significantly lower for multilingual tokenizers than for monolingual English tokenizers (\cref{fig:costs_training_multi}). 
In fact, \cref{fig:costs_training_de} and \cref{tab:train_costs} in the appendix demonstrate that the training costs can increase up to 68\% (comparing Multi-UNI-SP-50 to EN-UNI-SP-100 for German documents) for a given dataset.
Assuming that during training it is required to process a fixed set of documents (e.g., Wikipedia to learn specific facts) entirely and not only a given number of tokens, the choice of the tokenizer can significantly impact the computational costs for training on this corpus.

While we could observe large cost differences between multilingual and monolingual English tokenizers in the monolingual English setting, the difference in computational costs between multilingual and monolingual English tokenizers for processing English documents is marginal (\cref{fig:costs_training_en}).

\section{Correlation Between Intrinsic And Extrinsic Tokenizer Performance} \label{sec:corr}

\begin{figure}
        \includegraphics[width=0.5\textwidth]{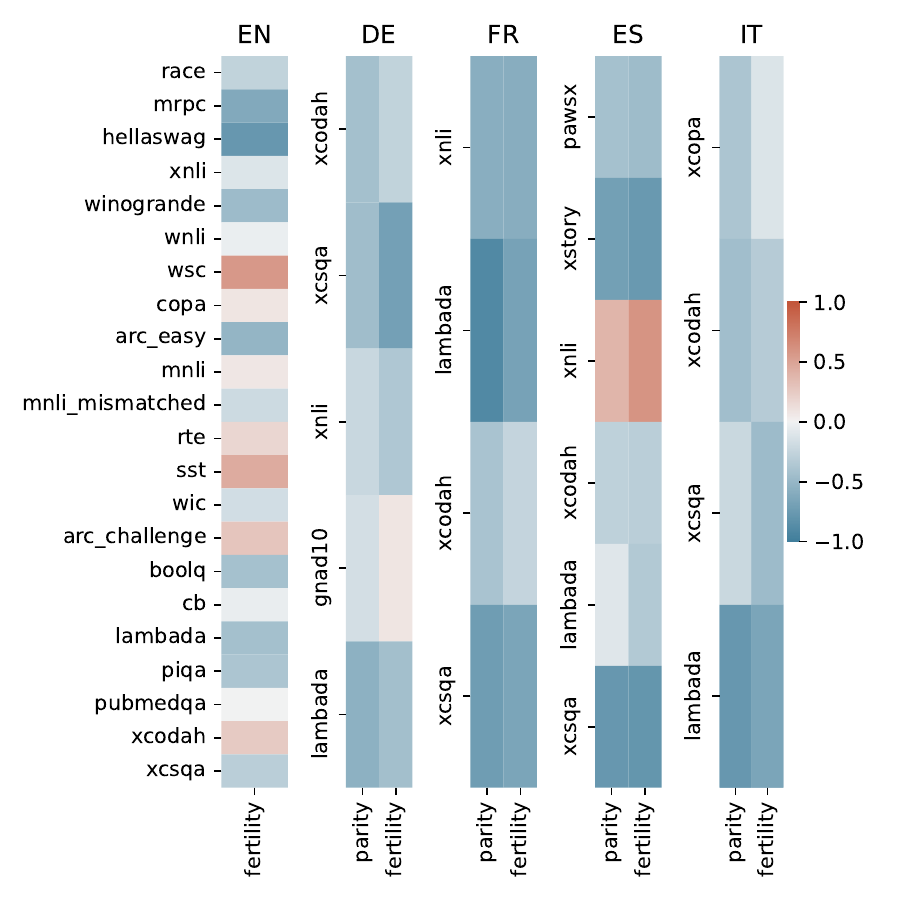}
        \caption{Spearman correlation of fertility/parity scores and downstream task performance for all five languages. 
        We evaluated monolingual models on English tasks (left), whereas our multilingual models are evaluated across all non-English tasks. Pearson and Kendall correlation metrics showed a very similar picture.}
        \label{fig:fertility_corr}
\end{figure}

This section investigates a possible predictive relationship of intrinsic tokenizer metrics (fertility and parity) to the extrinsic model downstream performance. 

As highlighted in the correlation heatmaps in \cref{fig:fertility_corr}, we find that there is no distinct correlation across all tasks and languages, demanding a more granular analysis.
While for non-English tasks, we mainly observe a correlation between low fertility and higher downstream performance, the non-English tasks yield seemingly random positive and negative correlations.
However, it should be noted that the number of multilingual tasks per language is much lower than for English and that for several multilingual tasks such as XSQA and LAMBADA, a similar correlation behaviour between the English tasks and their translated version can be observed.

Taking the fertility trends with varying vocabulary sizes (see \cref{fig:fertility}) into consideration, we hypothesize that fertility only correlates with downstream performance in certain language-specific vocabulary size limits. For the English language, the tokenizers already provide low, close-to-convergence fertility scores for vocabulary sizes of 33k tokens. While additional tokens yield only minute fertility improvements, we presume that they do not capture morphological segmentations and, thus, can harm downstream performance and significantly increase the computation costs (see \cref{subsec:costs}) in the end.

In contrast, for multilingual tokenizers, we observe significant fertility improvements with increasing vocabulary sizes. Due to the larger thesaurus induced by the additional languages, the tokenizer requires a larger vocabulary to allow a model to perform convincingly on all languages. Therefore, only within the non-convergence vocabulary range, we achieve a strong, negative correlation between fertility and downstream performance with varying vocabulary sizes.

In conclusion, intrinsic tokenizer metrics such as fertility and parity need to be taken with a grain of salt and supposedly are only predictive of downstream model performance in certain bounds. 
Low fertility scores might be regarded as a necessary criterion but not as a sufficient one.

\section{Conclusion \& Future Work}
This work represents a fundamental step to a better understanding of the impact of the tokenizer on the models' downstream performance. 
We have shown that training tokenizers with a balanced share across languages achieve comparable low fertility and parity scores across all languages, which has important implications.
Higher fertility results in up to 68\% more computational costs during training and prevents the model from learning long-range dependencies in limited context windows.

Furthermore, we highlight that the tokenizer choice can significantly impact the model's downstream performance. 
We could show that the BPE algorithm applies well to mono- and multilingual settings. 
For English, we show that a vocabulary size of 33k is sufficient, whereas multilingual models based on our five considered languages require a up to three times larger vocabulary size. 
Moreover, we could show that the SentencePiece library outperforms the Huggingface tokenizer library.

Finally, we could demonstrate that there is no clear correlation between intrinsic and extrinsic tokenizer performance, but the correlation is rather task-specific. 
A small fertility value might be a necessary condition for good downstream performance but not a sufficient one.

In the future, we aim to investigate tokenizers for a larger set of languages, including very diverse languages, and investigate the impact of alternative tokenization approaches such as SAGE~\cite{DBLP:conf/eacl/YehezkelP23} that focus on context information during tokenizer training.

\section{Limitations}

Despite the extensiveness of our work, it faces the following limitations.

Firstly, we did not perform hyper-parameter optimizations for each tokenizer. 
This was a deliberate choice to avoid additional computational costs, considering that training all 26 models only once required $\approx59.000$ GPU hours.

Secondly, we did not investigate the effect of different random seeds on the model performance for a given tokenizer due to the additional computational costs.
However, our results lay the foundation for future works that can further investigate the robustness of selected experiments.

Third, we did not investigate whether the results obtained could be extrapolated to larger model sizes, which we leave to future works. 
However, our finding that the BPE-SP-33 tokenizer is the best-performing tokenizer for the monolingual setting and the fact that this tokenizer has been used for training state-of-the-art models up to 65B~\cite{touvron2302llama} might indicate that our results also transfer to larger model sizes.

Finally, we did not provide results for a few-show setting since the metric of interest in the context of this work was the zero-shot downstream performance.
Because we wanted to investigate whether the tokenizer choice impacts the model's downstream performance, we argue that restricting on one of the widely applied metrics, i.e., the zero-shot setting, is sufficient to answer this research question.
One further advantage of focusing on the zero-shot scenario is that we do not introduce an additional variable represented by the choice of the few-shot examples. 
However, we encourage future works to investigate whether our results translate into the few-shot evaluation setting.

\section{Ethical And Broader Impact}

LLMs represent a disruptive technology that has received significant attention from the public and is widely used across societies speaking different languages.
Therefore, ensuring a democratization of the technology across people of different languages will represent an important value.
Our study highlights that neglecting multilingualism while training a tokenizer representing a core component required for training LLMs can cause severe disadvantages, such as increased training costs and decreased downstream performance, raising major ethical concerns.
Furthermore, the increased training costs translate into an increased carbon footprint, which has an environmental impact.
Our findings support an improved development and usage of this fundamental technology.

\section*{Acknowledgements}

This work was funded by the German Federal Ministry for Economic Affairs and Climate Action (BMWK) through the project OpenGPT-X (project no. 68GX21007D) as well as by the Federal Ministry of Education and Research of Germany and the state of North-Rhine Westphalia as part of the Lamarr-Institute for Machine, LAMARR22B and by the European Union’s Horizon 2020 research and innovation program under grant agreement No. 101135671 (TrustLLM) and 952215 (TAILOR). The authors gratefully acknowledge the Gauss Centre for Supercomputing e.V. (www.gauss-centre.eu) for funding this project by providing computing time on the GCS Supercomputer JUWELS at Jülich Supercomputing Centre (JSC) as well as the Center for Information Services and High Performance Computing [Zentrum für Informationsdienste und Hochleistungsrechnen (ZIH)] at TU Dresden for providing its facilities for high throughput calculations.

\bibliography{custom}

\appendix

\appendix
\section{Corpora}
\begin{table}
    \begin{tabular}{llr}
    \toprule
    Name & Language & \#Words   
\\
\midrule
Oscar& DE & 11.200.000.000\\
Oscar& ES & 11.200.000.000 \\
Oscar& EN & 11.200.000.000 \\
Oscar& IT & 11.200.000.000 \\
Oscar& FR & 11.200.000.000 \\
\midrule
Pile& DE  & 13.838.432\\
Pile& ES & 21.990.512 \\
Pile& EN & 4.334.313.669 \\
Pile& IT & 7.946.402 \\
Pile& FR & 15.857.811 \\
\midrule
RedPajama& DE  & 143.907.461\\
RedPajama& ES & 112.950.000 \\
RedPajama& EN & 4.663.646.781 \\
RedPajama& IT & 137.802.711 \\
RedPajama& FR & 139.749.147 \\
RedPajama& Code & 2.052.228.788 \\

\midrule
Misc& DE  & 600.844.912\\
Misc& ES & 186.934.269 \\
Misc& EN & 1.337.030.904 \\
Misc& IT & 19.810.753 \\
Misc& FR & 211.147.445 \\
\midrule
Total & & 70.000.000.000 \\
\bottomrule
\end{tabular}
\label{tab:multilingual_dataset}
\caption{Overview of the multilingual 70B words dataset with language, number of sampled words}
\end{table}

\begin{table}
    \begin{tabular}{llr}
    \toprule
    Name & Language & \#Words   
\\
\midrule
Oscar& EN & 56.000.000.000 \\
Pile& EN & 4.893.724.288 \\
RedPajama& EN & 5.308.974.750 \\
RedPajama& Code & 2.299.301.635 \\
Misc& EN & 1.497.999.327 \\
\bottomrule
Total & & 70.000.000.000 \\
\bottomrule
\end{tabular}
\label{tab:monolingual_dataset}
\caption{%
Overview of the English 70B words dataset with language, number of sampled words}
\end{table}

\label{appendix:corpora}
Our web documents in the corpora consist of Oscars\footnote{\url{https://oscar-project.org/}} \cite{abadji_ortiz_etal2021}, that were generated by the ungoliant pipeline\footnote{\url{https://github.com/oscar-project/ungoliant}} based on three Common Crawl WET Archives (2022-27, 2022-49 and 2023-14). 

The curated datasets consist of \textit{The Pile} \cite{pile2020}, \textit{RedPajama} \cite{together2023redpajama}, and single datasets that do not belong to a collection. From the Pile subcorpora, we selected: Phil Archive, PMC Abstracts, PMC Extracts, OpenWebText, NIH Exporterm, and Free Law Opinions V2. From RedPajama we use: ArXiv, Books, Github, StackExchange, and Wikipedia.

The remaining datasets are:

\begin{enumerate}
    \item All the News V2.0\footnote{\url{https://metatext.io/datasets/all-the-news-2.0}} is a corpus of newspaper 
    articles crawled from over 26 different publications 
    from January 2016 to April 1, 2020.
    \item Bundestag - Plenarprotokolle\footnote{\url{https://www.bundestag.de/dokumente/protokolle/plenarprotokolle}}
comprises transcripts of sessions of the German Bundestag.
    \item Bundesgerichtshof - Entscheidungen\footnote{\url{https://www.bundesgerichtshof.de/DE/Entscheidungen/entscheidungen_node.html}}
is a collection of decisions of the German Federal Court. 
    \item CoStEP\footnote{\url{https://pub.cl.uzh.ch/wiki/public/costep/start}}
is a cleaned-up and corrected version of the EuroParl corpus\cite{graen_batinic_etal2014}.
\cite{koehn2005}
    \item DCEP\footnote{\url{https://joint-research-centre.ec.europa.eu/language-technology-resources/dcep-digital-corpus-european-parliament_en}}
is a companion corpus to CoStEP, containing documents published by 
    the European Parliament.
\cite{hajlaoui_kolovratnik_etal2014}
\item DNB Dissertations\footnote{\url{https://www.dnb.de/DE/Professionell/Services/Dissonline/dissonline_node.html}} is a collection of dissertations from the 
    Deutsche Nationalbibliothek.
\item MAREC/IREC\footnote{\url{https://researchdata.tuwien.ac.at/records/2zx6e-5pr64}}: The MAtrixware REsearch Collection / The Information retrieval facility Research Collection is a patent corpus of over 19 million documents from the EP, WO, US, and JP patent offices. 
\item Medi-Notice\footnote{\url{https://pub.cl.uzh.ch/wiki/public/pacoco/medi-notice}}
is part of the
Zurich Parallel Corpus Collection. It is
a multilingual corpus
compiled from information leaflets for medications and pharmaceutical products 
published by the Swiss Agency for Therapeutic Products.\cite{graen_kew_etal2019}
\item Swiss Policy\footnote{\url{https://pub.cl.uzh.ch/wiki/public/pacoco/swiss_legislation_corpus}} contains documents of the
Swiss Legislation Corpus \cite{hoefler_piotrowski2011}
\item OpenSubtitles 2018\footnote{\url{https://opus.nlpl.eu/OpenSubtitles-v2018.php}}\footnote{\url{https://www.opensubtitles.org/de/index.cgi}} is a collection of translated movie subtitles.
\cite{lison_tiedemann2016}
\end{enumerate}

\section{Tokenizer}\label{appendix:tokenizer}
In our experiments, we focused on the \textit{Huggingface tokenizer} library \cite{Moi_HuggingFace_s_Tokenizers_2023}  and the \textit{SentencePiece} library \cite{kudo2018sentencepiece}. We use the standard settings of the SentencePiece library if not stated otherwise in \cref{tab:sentencepiece}. 
For the HuggingFace tokenizer library \cref{tab:huggingface} shows where we deviated from the standard values.

\begin{table}
\centering
\begin{tabular}{ll}
\toprule
Hyper-Parameter & Value(s)\\
\midrule
model\_type & Unigram $|$ BPE \\
vocab\_size & 33k $|$50k\\
 & 82k $|$ 100k \\
character\_coverage  & 0.9999 \\
split\_by\_number & True \\
allow\_whitespace\_only& True \\
add\_dummy\_prefix& True\\
user\_symbols\ & $$<s>,</s>,<pad>,$$ \\
 & $$<eod>, <ph\_1>,$$ \\
  & $$ \dots, <ph\_255>$$ \\

byte\_fallback & True \\
max\_sentence\_length & 4192 \\
normalization\_rule\_name & NFKC \\
train\_large\_corpus & True \\
remove\_extra\_whitespaces & False \\
split\_by\_whitespace & True \\
\bottomrule
    \end{tabular}
    \caption{Overview of the SentencePiece options that we used for the training of our tokenizers.}
    \label{tab:sentencepiece}
\end{table}

\begin{table}
\centering
\begin{tabular}{ll}
\toprule
Hyper-Parameter & Value(s)\\
\midrule
model\_type & BPE \\
vocab\_size & 33k $|$ 50k \\
 & 82k $|$ 100k \\

limit\_alphabet  & 512 \\
nfkc\_normalizer  & True \\
lowercase\_normalizer & False \\
strip\_accents\_normalizer  & True \\
pre\_tokenizer  & ByteLevel, Digits \\
\bottomrule
    \end{tabular}
    \caption{Overview of the Huggingface options that we used for the training of our tokenizers.}
    \label{tab:huggingface}
\end{table}

\section{LLM Architecture and Hyperparameters}

\label{appendix:architecture_params}
Regarding the training architecture of our 2.6B parameter models, we followed closely the architecture of GPT-3 \cite{brown2020language}. 
An overview of the used architecture details and hyperparameters is given in \cref{tab:hyperparams}.
\begin{table}
\begin{tabular}{ll}
    \toprule
    Hyper-Parameter & Value \\
\midrule
    \# Hidden Dimension & 2560 \\
    \# Layers & 32 \\
    \# Attention-Heads & 32 \\
    Sequence-Length & 2048 \\
    Optimizer & Adam \\
    Adam$-\beta_1$ & 0.9 \\
    Adam$-\beta_2$ & 0.9 \\
    Learning rate & 1.6e-4  \\ 
    Learning rate decay & Cosine  \\ 
    Precision & BF16 \\ 
    FlashAttention & 2.0 \\ 
    Position-Embeddings & Rotary \\ 
    \bottomrule
    \end{tabular}
    \caption{Overview of the LLM hyperparameters that we used for the training.}
    \label{tab:hyperparams}
\end{table}

For training the models, we used a fork of Megatron-LM\url{https://github.com/NVIDIA/Megatron-LM}.

\section{Intrinsic Tokenizer Evaluation}\label{appendix:intrinsic}
\begin{figure*}
     \centering
     \begin{subfigure}[t]{0.49\textwidth}
         \centering
         \includegraphics[width=\textwidth]{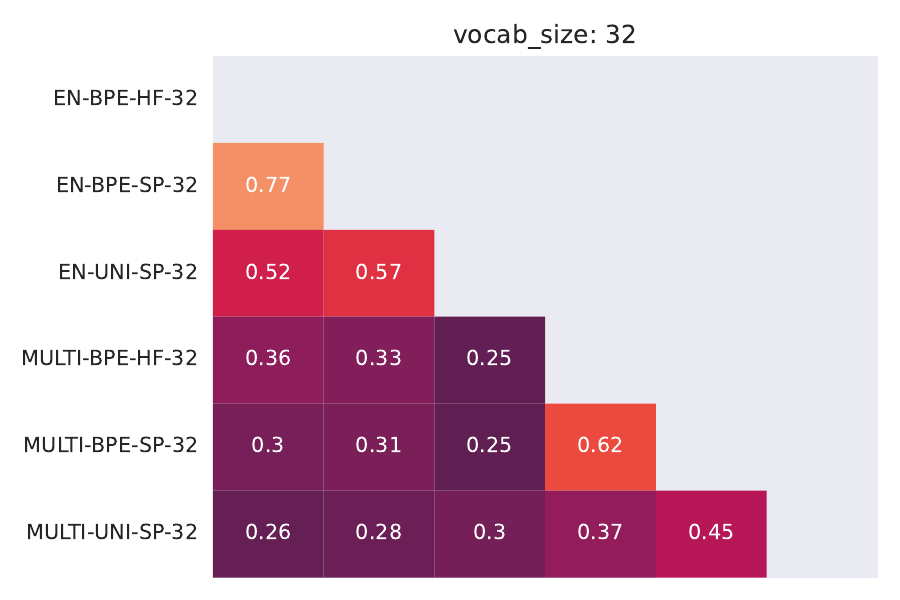}
     \end{subfigure}
        \begin{subfigure}[t]{0.49\textwidth}
         \centering
         \includegraphics[width=\textwidth]{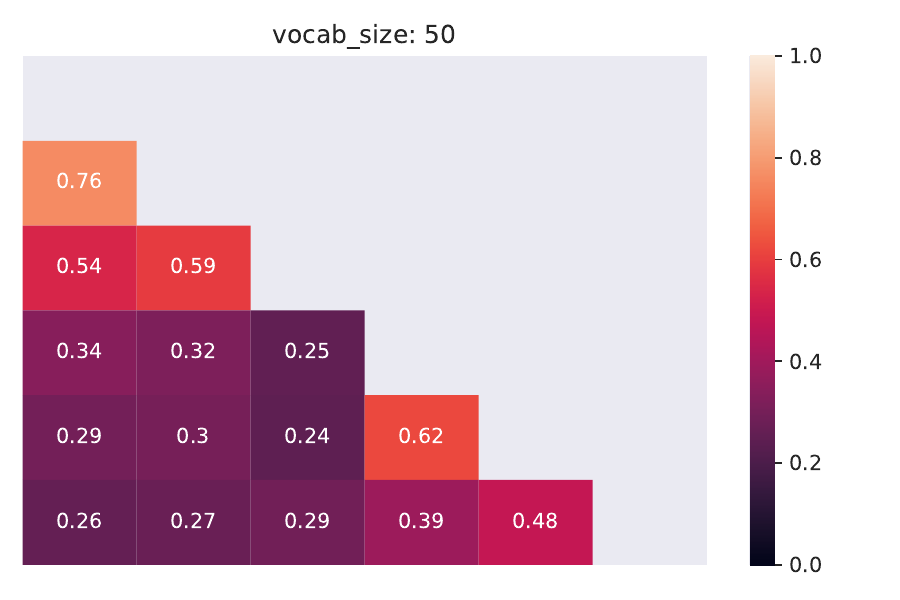}
     \end{subfigure}\\
     \begin{subfigure}[b]{0.49\textwidth}
         \centering
         \includegraphics[width=\textwidth]{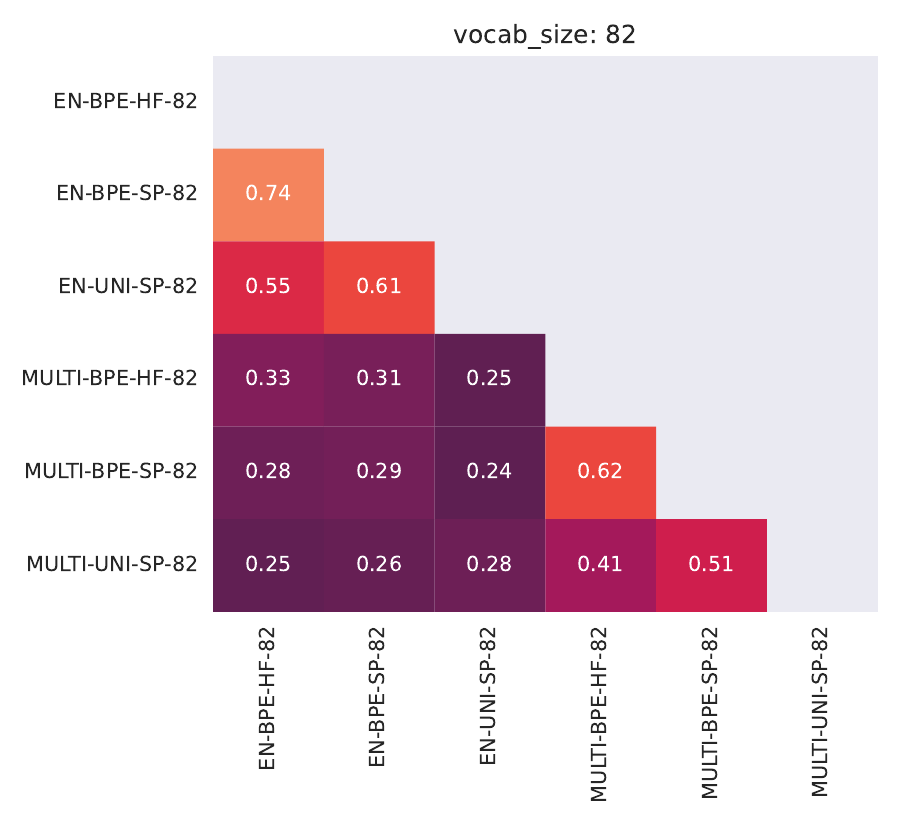}
     \end{subfigure}
        \begin{subfigure}[b]{0.49\textwidth}
         \centering
         \includegraphics[width=\textwidth]{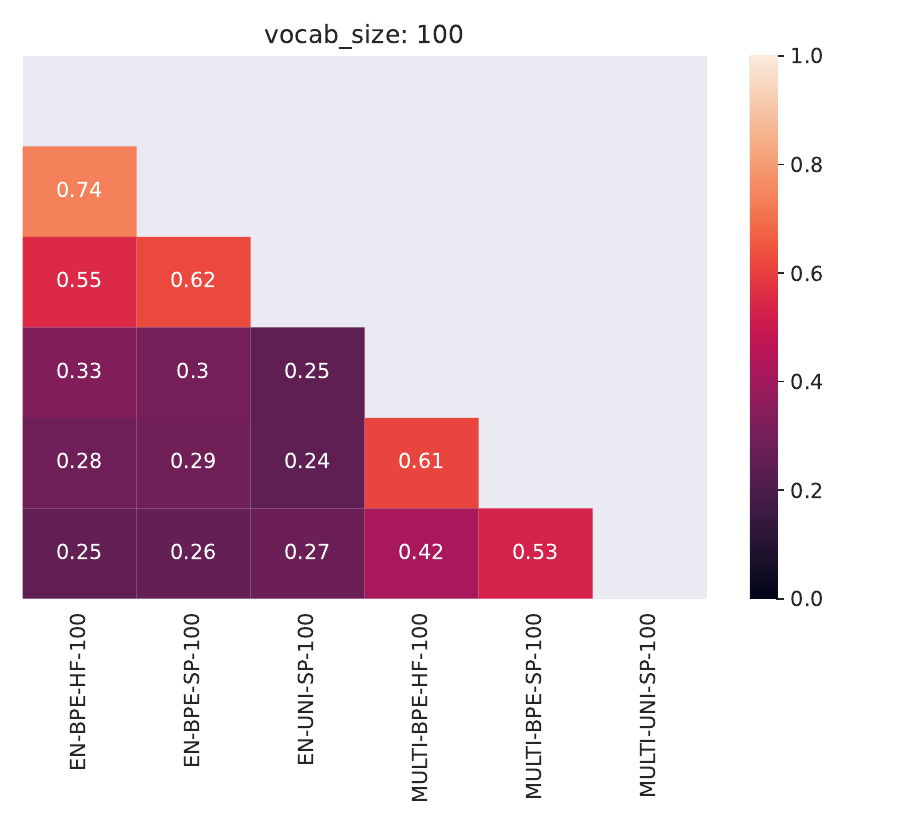}
     \end{subfigure}
        \caption{Vocabulary overlap between the examined tokenizers}
        \label{fig:overlap_across_tokenizers}
\end{figure*}

Besides studying the overlap of the same algorithm on the same thesaurus, we were also interested in vocabulary overlaps across algorithms and thesauruses see \cref{fig:overlap_across_tokenizers}. What we can observe is that multilingual vocabulary and English vocabulary have a rather small overlap between 24\% and 34\% that remains similar across increasing vocabulary sizes. Across algorithms, we can see that Unigram and BPE of SentencePiece have a slightly higher overlap than Unigram of SentencePiece and BPE of Huggingface. We think this might be due to library-specific preprocessing steps and more similar hyperparameters.

\subsection{Computational Costs Per Word During Training}

\cref{tab:train_costs} shows the average computational training costs for processing a word during the forward and backward pass.

\begin{table}
\centering
\begin{tabular}{llccc}
    \toprule
    & Model & Non-English & English & German \\
    \midrule
    & GPT-2-50 & 3.87 & 2.58 & 4.59 \\
    \midrule
    \multirow{12}{*}{\rotatebox{90}{EN}} 
    & BPE-HF-33 & 3.8 & \textbf{2.32} & 4.52 \\
    & BPE-HF-50 & 3.79 & 2.38 & 4.45 \\
    & BPE-HF-82 & 3.88 & 2.55 & 4.51 \\
    & BPE-HF-100 & 3.96 & 2.67 & 4.58 \\
    & BPE-SP-33 & 3.86 & 2.37 & 4.66 \\
    & BPE-SP-50 & 3.89 & 2.42 & 4.68 \\
    & BPE-SP-82 & 4.02 & 2.59 & 4.78 \\
    & BPE-SP-100 & 4.11 & 2.71 & 4.84 \\
    & UNI-SP-32 & 4.01 & 2.36 & 4.73 \\
    & UNI-SP-50 & 4.02 & 2.42 & 4.75 \\
    & UNI-SP-82 & 4.12 & 2.59 & 4.83 \\
    & UNI-SP-100 & 4.21 & 2.71 & 4.88 \\
    \midrule
    \multirow{12}{*}{\rotatebox{90}{MULTI}} 
    & BPE-HF-33 & 2.71 & 2.46 & 3.04 \\
    & BPE-HF-50 & 2.7 & 2.5 & 3.01 \\
    & BPE-HF-82 & 2.8 & 2.65 & 3.09 \\
    & BPE-HF-100 & 2.88 & 2.76 & 3.17 \\
    & BPE-SP-33 & 2.68 & 2.55 & 2.99 \\
    & BPE-SP-50 & 2.67 & 2.57 & 2.95 \\
    & BPE-SP-82 & 2.76 & 2.72 & 3.03 \\
    & BPE-SP-100 & 2.85 & 2.82 & 3.1 \\
    & UNI-SP-33 & 2.68 & 2.55 & 2.94 \\
    & UNI-SP-50 & \textbf{2.66} & 2.58 & \textbf{2.91} \\
    & UNI-SP-82 & 2.76 & 2.73 & 2.99 \\
    & UNI-SP-100 & 2.84 & 2.83 & 3.07 \\
    \bottomrule
\end{tabular}
\caption{Computational training costs per word (GFLOPs) for different tokenizers.}
\label{tab:train_costs}
\end{table}

\section{Infrastructure \& Computational Costs}

We trained each of our 26 2.6B parameter models on NVIDIA A100 GPUs, and the training of each model took up to 2304 GPU hours.
Therefore, the total training costs amounted to $\approx59.000$ GPU hours.

\end{document}